\documentclass[letterpaper, 10 pt, journal, twoside]{ieeetran}
\IEEEoverridecommandlockouts  

\usepackage{bm}
\usepackage{tikz}
\usepackage{xargs}
\usepackage[hyphens]{url}
\usepackage{times}
\usepackage{array}
\usepackage{xcolor}
\usepackage{siunitx}
\usepackage{amssymb}
\usepackage{amsmath}
\usepackage{flushend}
\usepackage{graphics}
\usepackage{graphicx}
\usepackage{hyperref}
\usepackage{multicol}
\usepackage{verbatim}
\usepackage{multirow}
\usepackage{booktabs}
\usepackage{pgfplots}
\usepackage{algorithm}
\usepackage{glossaries}
\usepackage{algorithmic}
\usepackage[nolist]{acronym} \newacro{vs}[VS]{Visual Servoing}
\newacro{pulp}[PULP]{Parallel Ultra-Low Power}
\newacro{dof}[DoF]{Degrees of Freedom}
\newacro{nn}[NN]{Neural Network}
\newacro{cnn}[CNN]{Convolutional Neural Network}
\newacro{fcn}[FCN]{Fully Convolutional Network}
\newacro{fc}[FC]{Fully-Connected}
\newacro{ml}[ML]{Machine Learning}
\newacro{ssl}[SSL]{Self-Supervised Learning}
\newacro{ros}[ROS]{Robot Operating System}
\newacro{dl}[DL]{Deep Learning}
\newacro{mav}[MAV]{Micro Aerial Vehicle}
\newacro{uwb}[UWB]{Ultra-Wide Band}
\newacro{rss}[RSSI]{Received Signal Strength Intensity}
\newacro{ekf}[EKF]{Extended Kalman filter}
\newacro{auc}[AUC]{Area Under the Receiver Operating Characteristic Curve}
\newacro{soa}[SoA]{State-of-the-art}
\newacro{soc}[SoC]{System-on-Chip}

\newacro{ledp}[LED-P]{LED state prediction Pretext}
\newacro{decro}[EDNN]{Efficient Deep Neural Networks}
\newacro{bas}[BAS]{Baseline}
\newacro{ub}[UB]{Upper Bound}
\newacro{ae}[AE-P]{Autoencoding Pretext}
\newacro{clip}[CLIP]{Contrastive Language-Image Pre-training}
\usetikzlibrary{arrows.meta}
\usepgfplotslibrary{groupplots}
\usepgfplotslibrary{fillbetween}
\pgfplotsset{compat=1.4}
\graphicspath{{fig/}}
\setacronymstyle{long-short}
\DeclareMathAlphabet{\mathpzc}{OT1}{pzc}{m}{it}

\newcommand{\mdiff}[1]{#1}

\title{Self-Supervised Learning of\\Visual Robot Localization\\Using LED State Prediction as a Pretext Task
}

\author{Mirko Nava$^{1}$, Nicholas Carlotti$^{1}$, Luca Crupi$^{1}$, Daniele Palossi$^{12}$, and Alessandro Giusti$^{1}$
\thanks{Manuscript received: September 13, 2023; Revised November 9, 2023; Accepted January 26, 2024. This paper was recommended for publication by Editor X. Liu upon evaluation of the Associate Editor and Reviewers' comments.
This work was supported by the Swiss National Science Foundation, grant number 213074.} 
\thanks{$^{1}$M. Nava, N. Carlotti, L. Crupi, D. Palossi, and A. Giusti are with the Dalle Molle Institute for Artificial Intelligence (IDSIA), USI-SUPSI, Lugano, 6962, Switzerland {\tt\footnotesize mirko@idsia.ch}}%
\thanks{$^{2}$D. Palossi is also with the Integrated Systems Laboratory (IIS), ETH
Z\"{u}rich, Z\"{u}rich, 8092, Switzerland}
\thanks{Digital Object Identifier (DOI): see top of this page.}%
}

\begin{document}

\markboth{IEEE Robotics and Automation Letters. Preprint Version. Accepted January, 2024}
{Nava \MakeLowercase{\textit{et al.}}: Self-Supervised Learning of Visual Robot Localization Using LED State Prediction as a Pretext Task}

\maketitle


\begin{abstract}
We propose a novel self-supervised approach for learning to visually localize robots equipped with controllable LEDs. 
We rely on a few training samples labeled with position ground truth and many training samples in which only the LED state is known, whose collection is cheap.
We show that using LED state prediction as a pretext task significantly helps to learn the visual localization end task.
The resulting model does not require knowledge of LED states during inference.

We instantiate the approach to visual relative localization of nano-quadrotors:
experimental results show that using our pretext task significantly improves localization accuracy (from 68.3\% to 76.2\%) and outperforms alternative strategies, such as a supervised baseline, model pre-training, and an autoencoding pretext task.
We deploy our model aboard a 27-g Crazyflie nano-drone, running at 21 fps, in a position-tracking task of a peer nano-drone.
Our approach, relying on position labels for only 300 images, yields a mean tracking error of 4.2 cm versus 11.9 cm of a supervised baseline model trained without our pretext task.
Videos and code of the proposed approach are available at \url{https://github.com/idsia-robotics/leds-as-pretext}.
\end{abstract}


\section{Introduction}\label{sec:intro}

\IEEEPARstart{T}{he} ability to estimate the position of a target robot in a video feed is crucial for many robotics tasks~\cite{taha2019machine, pavliv2021tracking, ciani2023cyber}.
\ac{soa} approaches use deep learning techniques based on \acp{cnn}~\cite{li2022self}: given a camera frame, they segment the target robot, regress the coordinates of its bounding box or its position in the image.
Training these approaches to handle new robots or environments requires extensive labeled datasets, which are time-consuming and expensive to acquire, often relying on specialized hardware, e.g., motion tracking systems, to generate ground truth labels.

This article presents an approach to drastically reduce the labeled data required to train such models, building upon recent results in Self-Supervised Learning~\cite{jing2020self}.
In the robotics literature (see Section~\ref{sec:rw}), the term Self-Supervised denotes two distinct paradigms.
In the first, a robot system autonomously generates labeled data for the task of interest, named \emph{end task}, and is trained in a standard supervised way.
This paradigm has been used in robotics since the mid 2000s~\cite{dahlkamp2006self,lookingbill2006reverse,hadsell2009learning,zeng2017multi}.
As a recent example, Li et al.~\cite{li2022self} use nano-drones equipped with \ac{soa} algorithms to automatically acquire camera frames and the corresponding relative location of the target drone.
%
In the second paradigm, a robot system autonomously generates abundant labeled data for a \emph{pretext task}:
%
the pretext task requires similar perception skills as the end task while relying on cheaper ground truth that is easier or free to collect.
Then, a model is trained to solve both tasks simultaneously using an additional dataset containing only few labels for the end task.
Despite not being useful during deployment, the pretext task forces the model to learn meaningful features, boosting the performance on the end task.
The paradigm is widely successful in the deep learning literature~\cite{jing2020self} and has only recently been adopted for perception applications~\cite{nava2022learning, radosavovic2023real}.

This article introduces a novel approach based on the second paradigm, tailored to robotics applications, and suitable for deployment on resource-constrained platforms.
%
Our \textbf{contribution}, presented in Section~\ref{sec:model}, is the use of target robot LED state prediction (\textsc{on} or \textsc{off}) as a pretext task to improve the learning process of a visual localization end task.
By learning to predict the state of the LEDs aboard, the model learns features that are also useful to localize the target robot.
The idea is compelling because most robot platforms feature controllable LEDs:
during data collection, the target robot blinks its LEDs and radio-broadcast their state;
at the same time, another robot automatically collects images annotated with LED state ground truth.
%
%

\begin{figure*}[th]
    \centering
    \includegraphics[width=0.77\linewidth]{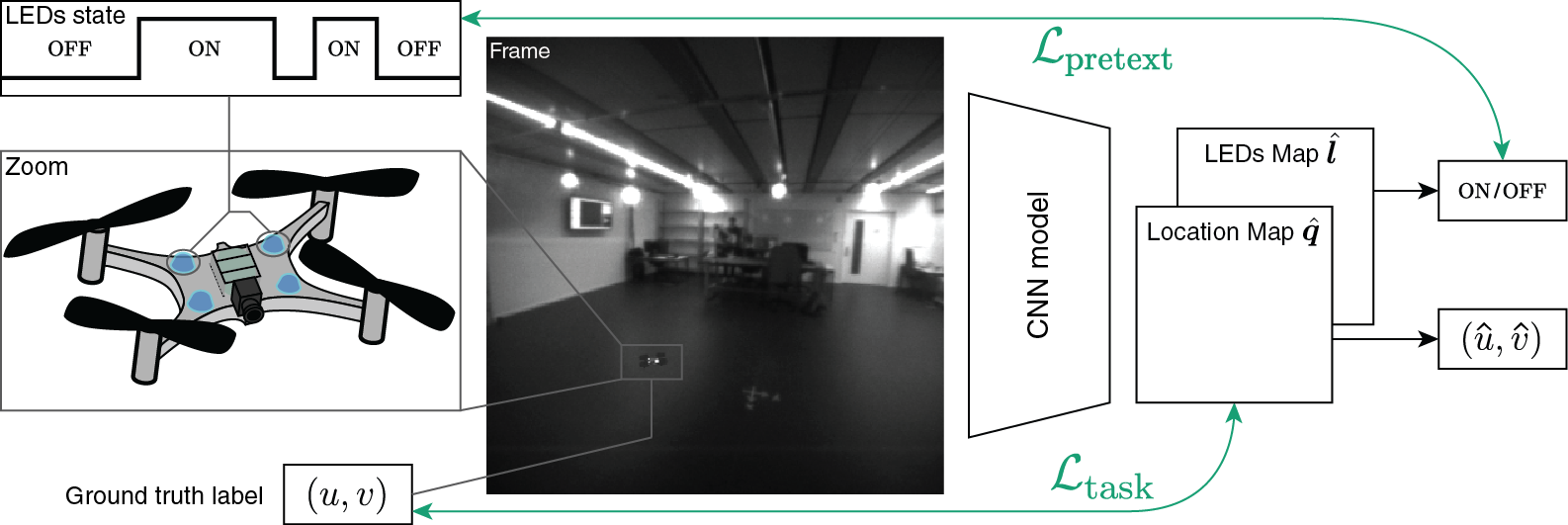}%
    \caption{A fully convolutional network model is trained to predict the drone position in the current frame by minimizing a loss ${\cal L}_\text{task}$ defined on a small labeled dataset ${\cal T}_l$ (bottom), and the state of the four drone LEDs, by minimizing ${\cal L}_\text{pretext}$ defined on a large dataset ${\cal T}_l \cup {\cal T}_u$ (top).}
    \label{fig:approach}
\end{figure*}

\mdiff{We instantiate this general idea to a specific, challenging end task: predict the image-space position of a target nano-drone given a low-resolution, low-dynamic-range image acquired by the camera of a peer nano-drone, as shown in Figure~\ref{fig:approach}}.
A \ac{fcn} model~\cite{long2015fully} simultaneously learns to solve pretext and end tasks using the dataset described in Section~\ref{sec:setup}, containing only few samples labeled with the drone's location.
We provide detailed comparisons and an ablation study on the Bitcraze Crazyflie 2.1\footnote{\href{https://www.bitcraze.io/products/Crazyflie-2-1}{https://www.bitcraze.io/products/Crazyflie-2-1}} nano-drone in Section~\ref{sec:results}.
Results show the proposed pretext task to significantly improve performance over a supervised baseline, different pre-training strategies, and an autoencoding pretext task.
The model generalizes well to unseen environments, and is capable of localizing multiple drones simultaneously.
Finally, we deploy the model aboard the target platform to complete a vision-based position tracking task.
Conclusions are drawn in Section~\ref{sec:conclusions}.

\section{Related Work}\label{sec:rw}
\subsection{Relative Visual Localization of Drones}

Drone-to-drone relative localization approaches rely on various sensors, including microphones, infra-red sensors, \ac{uwb}, color and depth cameras.
In particular, microphones can be used for localization, integrating distance estimates from a drone beacon emitting a specific sound~\cite{basiri2014audio}.
Multiple infra-red sensors with known geometry allow the triangulation of a drone equipped with infra-red emitters~\cite{roberts20123D}.
Camera-based approaches rely on visual fiducial markers such as circles printed on paper~\cite{saska2017system}, light-emitting markers~\cite{dias2016onboard}, by detecting the drone in depth images with handcrafted~\cite{vrba2019onboard}, or learned~\cite{carrio2020onboard} models.
In our work we use monocular grayscale images as the model's input.
LEDs, which come already integrated with the adopted platform, are exclusively used to generate data for the self-supervised pretext task and are not used during inference. 

%
%
\mdiff{\ac{uwb} is a radio communication technology recently adopted for localization tasks~\cite{guler2020peer, nguyen2019distance, coppola2018board, van2020board, xu2020decentralized}, enabling communication between multiple robots and providing a distance measurement through the \ac{rss}.
\ac{rss} measures the amount of radio signal received from a source and is used to derive its distance.
Using three non-collinear \ac{uwb} sensors enables the triangulation of robots~\cite{guler2020peer}.
A single sensor requires more complex approaches, such as integrating distance measurements from \ac{uwb} beacon drones moving in a pattern~\cite{nguyen2019distance}.
Communication is used during localization to combine distance measurements with broadcasted state-estimates~\cite{coppola2018board, van2020board} and optimizing a camera-based initial guess~\cite{xu2020decentralized}.

In contrast, our approach does not require specialized hardware that supports the communication, and does not assume the target drone to share information with the observer drone during inference.}

\subsection{Self-Supervised Relative Drone Localization}

Self-supervised approaches proposed for object localization tasks~\cite{zeng2017multi,xiang2018posecnn,deng2020self,nava2021uncertainty} can, in principle, be used to localize drones.
Objects are localized by fitting their known 3D model onto a monocular image~\cite{xiang2018posecnn} or onto a point-cloud obtained by segmenting multiple RGB-D images~\cite{zeng2017multi}.
Other approaches do not require knowledge of the 3D model of the objects of interest. 
Instead, they learn by using a pre-trained model and moving the object to generate more training data~\cite{deng2020self} or by combining state estimates with sparse trusted information, e.g., that coming from a fiducial marker~\cite{nava2021uncertainty}.

\mdiff{Self-supervised relative drone localization approaches learn a model with limited access to labeled data, using \ac{uwb} to provide ground truth ~\cite{li2022self}, or a stereo microphone for an audio-based pretext task~\cite{nava2022learning}.
In detail, Li et al.~\cite{li2022self} pre-train a purely visual estimator using synthetic data, then fine-tune it using a small labeled dataset generated autonomously from \ac{uwb} nodes~\cite{van2020board}.
In contrast, our approach introduces a pretext task defined on images with no ground truth for the target position:
it is based solely on LED state estimation, and does not require additional hardware besides controllable LEDs -- which are present on most robot platforms.}

We explored cross-modal self-supervised learning of visual quadrotor localization in recent work~\cite{nava2022learning}, using images acquired by a ground robot equipped with a stereo microphone.
The pretext task consists of predicting features (intensity in various frequency bands) of the perceived sound of a quadrotor, given an image.
By solving this pretext task, the model is forced to learn features of the perceived sound that, in turn, are
informative of the drone's location.

\mdiff{The present work proposes a more general pretext task that does not rely on additional sensors, such as a microphone, and is suitable for applications with limited power budget. 
The only requirement is that the target robot is able to vary its appearance for the observer: in the absence of controllable LEDs, which are the most straightforward and convenient way to achieve this, one may rely on any other actuator that affects the robot appearance, e.g., raising a limb.}

\section{LED State Prediction as a Pretext Task}\label{sec:model}

We consider visual robot-to-robot localization problems, in which an observer robot has to predict the position of a target robot on the image plane.
The observer robot takes a monocular image from its forward-looking camera and predicts the position of the target robot visible in the image.
Additionally, we require the target robot to be equipped with controllable LEDs.

\mdiff{We collect tuples consisting of
$
    \bigl\{ \langle \bm{i}_j, \bm{p}_j, l_j \rangle \bigr\}_{j=1}^N
$ where $\bm{i} \in \mathbf{R}^{whc}$ denotes a camera frame of $w \times h$ pixels and $c$ channels,
$\bm{p} \in \mathbf{R}^2$ the image-space position of the robot,
and $l$ the shared state of the four robot LEDs, which can be either all \textsc{off}, represented with a 0, or all \textsc{on}, represented with a 1.

In the following, we call samples \emph{labeled} when the drone's position $\bm{p}$ is known or \emph{unlabeled} otherwise.
We denote the set containing the (possibly small) amount of labeled samples with ${\cal T}_\ell$ and the unlabeled set with ${\cal T}_u$.
We also collected a separate labeled set $\mathcal{Q}$ that serves as a testing set and on which we compute performance metrics.

We learn a \ac{nn} model $\bm{m}$ that, given a monocular image, predicts two maps: the location map $\hat{\bm{q}}$ containing likely drone locations, and the LED state map $\hat{\bm{l}}$ the probability of seeing a drone with its LEDs on,
$
     \left( \hat{\bm{q}}, \hat{\bm{l}} \right) = \bm{m} \left( \bm{i} | \bm{\theta} \right)
$, where $\bm{\theta}$ is the set of trainable network weights.
Specifically, we consider a Fully Convolutional Network (FCN) architecture consisting solely of convolutional layers and whose output consists of two maps.}
Using a map to represent the drone's location has two advantages compared to using the drone's coordinates~\cite{bonato2023ultra}.
First, it allows one to handle images with zero, one or more visible drones~\cite{li2022self}.
Second, it enforces an inductive bias by limiting the receptive field of each cell of the output map;
in fact, we expect the target drone to cover a small portion of the input image~\cite{rozantsev2016detecting}.

A ground truth location map $\bm{q} \in \left[ 0, 1 \right] ^{wh}$ of $w \times h$ cells is generated from the robot's position $\bm{p} = ( u, v )$:
we start with a map filled with zeros and place a circle of radius $r = 4$ pixels centered in $\bm{p}$, filled with ones and with a soft-edge transitioning to zero.
%

We train $\bm{m}$ by optimizing the weights $\bm{\theta}$ through gradient descent steps, minimizing the loss function ${\cal L}$.
The loss, in turn, is defined as the weighted sum of two terms:
the first term ${\cal L}_\text{task}$ consists of a regression loss computed on the labeled training set ${\cal T}_\ell$, whose aim is to learn the robot localization task, and defined as
\begin{equation}
    {\cal L}_\text{task} = \frac{1}{| {\cal T}_\ell |}
    \sum_{i=1}^{| {\cal T}_\ell |}
    \operatorname{mean} \left( \left| \hat{\bm{q}}_i - \bm{q}_i \right|^2 \right)
    \label{eq:loss_task}
\end{equation}
where $\operatorname{mean}$ is the average of the map cells.
The second term ${\cal L}_\text{pretext}$ consists of a classification loss defined on the union of the training sets ${\cal T}_\ell \cup {\cal T}_u$, learning the LED state prediction task, and defined as
\begin{equation} 
    {\cal L}_\text{pretext} = \frac{1}{| {\cal T}_u  \cup {\cal T}_\ell |} \sum_{i=1}^{| {\cal T}_u \cup {\cal T}_\ell |} \operatorname{BCE} \left( \operatorname{mean} \left( \hat{\bm{l}} \right), l \right)
    \label{eq:loss_pretext}
\end{equation}
where $\operatorname{BCE}$ is the binary cross-entropy.
To obtain the scalar $\hat{l}$ representing the probability of the drone's LEDs being on, we compute the average of the LED state map $\hat{\bm{l}}$\footnote{\mdiff{One may obtain $\hat{l}$ as the average of $\hat{\bm{l}}$ weighted by the map $\hat{\bm{q}}$; in our experiments, this resulted in less stable training and a lower performance.}}.

The complete loss function is ${\cal L} = ( 1 - \lambda )\,{\cal L}_\text{task} + \lambda\,{\cal L}_\text{pretext}$,
where $\lambda \in [0, 1]$ controls the tradeoff between the two loss terms during training.
In \eqref{eq:loss_task} and \eqref{eq:loss_pretext}, each loss is weighted by the reciprocal of the dataset size on which it operates,
ensuring that the impact of each loss during training is comparable when working on differently-sized datasets.

\section{Experimental Setup}\label{sec:setup}

In the following, we instantiate the presented approach to the challenging task of drone-to-drone localization, as shown in Figure~\ref{fig:platform}.
This task represents the broader set of image-based robot localization tasks, as many mobile robots feature cameras and controllable LEDs.
Among platforms to which our approach is applicable, we specifically selected nano-drones for our experiments:
they are difficult to localize due to their small dimensions and complex shape.
%
Additionally, they have constrained resources: the camera is low resolution, low dynamic range, and has a limited field of view; the onboard microprocessor imposes limitations on the breadth and depth of the \ac{nn}, especially for real-time applications.

\subsection{Robot Platform}

\begin{figure*}[t]
    \centering
    \includegraphics[width=\linewidth]{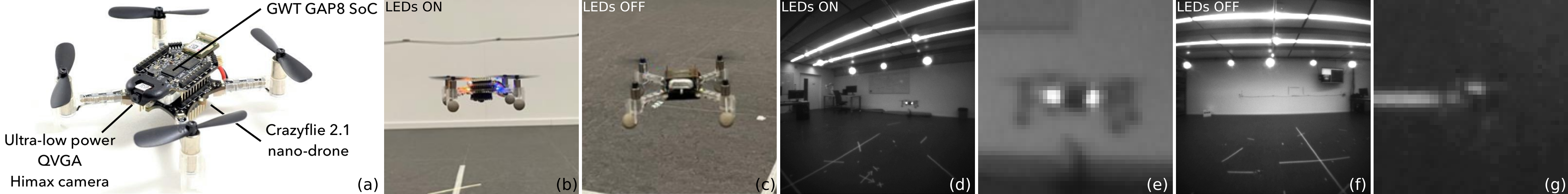}%
    \caption{The palm-sized Bitcraze Crazyflie 2.1 nano-drone platform (10 cm in diameter). (a) The drone's hardware and its four controllable LEDs; (b, c) high-resolution pictures of the flying drone; (d, f) samples from our dataset; (e, g) zoom-in on the drone using the model's receptive field ($45 \times 45$ pixels).}
    \label{fig:platform}
\end{figure*}

The platform of choice is the Bitcraze Crazyflie 2.1, a nano-drone measuring \SI{10}{cm} in diameter and weighing only \SI{27}{g}, extended by the Ai-deck companion board, see Figure~\ref{fig:platform}.
The Ai-deck provides a forward-looking monocular camera, acquiring $320 \times 320$ pixels grayscale images, and a GWT GAP8 \ac{pulp} \ac{soc}~\cite{palossi2019open} extending the basic computational capabilities, i.e., state estimation and low-level control, offered by the STM32 microcontroller available on the nano-drone.
\mdiff{We employ the GAP8 \ac{soc} to boost the execution of NNs, which require integer quantization to  exploit its 8-core general-purpose cluster due to the lack of floating point support.
%
%
Additionally, the drone features on its body four controllable LEDs, which we exploit to define the pretext task.}

We consider a scenario in which two identical Crazyflie drones fly in the environment: one drone takes the observer role, acquiring camera frames in which the other drone (target) is visible.

\subsection{Datasets}

Our experimental validation is based on Nano2nano\footnote{\mdiff{\href{https://github.com/idsia-robotics/drone2drone\_dataset}{https://github.com/idsia-robotics/drone2drone\_dataset}}},
a dataset collected in a $10 \times 10$m lab equipped with a motion-tracking system and consisting of 72 different sequences.
For each sequence (average length of 210 seconds and 830 frames), the target Crazyflie flies a pseudo-random trajectory with the four controllable LEDs switched either \textsc{on} or \textsc{off}.
\mdiff{
At the same time, the observer drone continuously moves to increase the variability of represented backgrounds.
To further increase data variability, in each sequence the camera exposure setting is set to one of three possible values.
The trajectory is computed so as to keep the target in the camera view and cover the image space as uniformly as possible, with distances ranging from 0.2 meters to 2 meters.
Each frame is labeled with the target pose relative to the observer's camera, its position in the image, and the state of the LEDs.}

Half of the 72 sequences are used as the testing set ${\cal Q}$ (30k samples).
Data from remaining 36 sequences is partitioned into the labeled training set ${\cal T}_\ell$ (1k samples) and the unlabeled training set ${\cal T}_u$ (29k samples).
\mdiff{For an approach described in Section~\ref{sec:setup:strategies}, we employ the synthetic training dataset $\mathcal{T}_{s}$ proposed in~\cite[Section IV]{li2022self} consisting of 800 random-background images depicting the drone in a random pose.
Images are converted to grayscale and padded with a solid random gray value to match size and channels of our data.}

Additional generalization experiments are reported in Section~\ref{sec:results:generalization} and shown in the supplementary video;
these experiments use data recorded in different rooms, without ground truth for the drone location.

\subsection{Alternative Strategies}\label{sec:setup:strategies}

We assess the validity of our approach, named \ac{ledp}, against various alternatives.
First, we consider a naive model (DUMMY) that always predicts the mean position on the labeled training set $\mathcal{T}_\ell$.  
The \ac{bas} strategy involves training using only $\mathcal{L}_\text{task}$ (achieved with $\lambda\,=\,0$) on the labeled training set $\mathcal{T}_\ell$.
The \ac{ub} strategy is used to estimate the maximum achievable performance.
It minimizes only $\mathcal{L}_\text{task}$, assuming to have access to ground truth position labels for both $\mathcal{T}_\ell$ and $\mathcal{T}_u$, representing a fully-supervised scenario where ground truth is cheap and abundant.

We also consider alternative strategies: \ac{ae}, \ac{clip} and \ac{decro}.

Undercomplete autoencoders are a frequently-adopted strategy for taking advantage of unlabeled data, defining an image-reconstruction pretext task~\cite{jing2020self}.
The intuition is that by learning to compress and decompress an image, autoencoders learn a high-level representation that can be useful to solve the end task.
In \ac{ae}, we train an autoencoder on $\mathcal{T}_\ell \cup \mathcal{T}_u$ by minimizing the MS-SSIM~\cite{wang2003multiscale} between input and reconstructed images; then, an additional \ac{nn} head learns the localization task using $\mathcal{L}_\text{task}$ on $\mathcal{T}_\ell$, taking as input features computed by the autoencoder's bottleneck.

\ac{clip} is a powerful bi-modal feature extractor trained to minimize the distance of embeddings between an image and its caption~\cite{radford2021learning}.
The learned image encoder is shown to outperform supervised models in many zero- and few-shot tasks.
In \ac{clip}, we take the features extracted from the pre-trained image encoder and pass them to a \ac{nn} head trained for the localization task using $\mathcal{L}_\text{task}$ on $\mathcal{T}_\ell$.

In \ac{decro}, we consider supervised pre-training on synthetic images, as described in~\cite{li2022self}, to cheaply generate labeled data.
The strategy consists first in training using the synthetic dataset $\mathcal{T}_s$ on the localization task with a focal loss~\cite{lin2017focal}, and then fine-tune the \ac{nn} parameters using $\mathcal{L}_\text{task}$ on $\mathcal{T}_\ell$.

\subsection{Network Architectures and Training}

\ac{bas}, \ac{ub}, \ac{decro} and \ac{ledp} share the same tiny \ac{fcn}~\cite{long2015fully} architecture consisting of nine convolution blocks, in order: two blocks with 8 channels, $2\times$ max-pooling, three with 16, $2\times$ max-pooling, three with 32, $2\times$ max-pooling and one with 2 channels as the output, totalling 22.1k trainable parameters.
Convolution blocks consist of a convolution layer, batch normalization and ReLU activation.
The model's input is a grayscale image of $320 \times 320$ pixels, normalized between zero and one.
The model produces two maps of $40 \times 40$ cells, each cell with a $45 \times 45$ pixels receptive field, as illustrated in (e, g) of Figure~\ref{fig:platform}.
Cells of the first map denote drone presence in the corresponding area of the input image\footnote{Map cells are independent one another and assume values in the range from 0 (no drone) to 1 (drone present in given cell).},
while cells of the second denote whether the drone has its LEDs on (1.0) or off (0.0);
cells with no visible drone are expected to have a value close to 0.5.

\ac{ae} uses an encoder with four convolution blocks with 4, 8, 16 and 32 channels, interleaved by $2\times$ max-pooling layers, and terminating with the \ac{fc} bottleneck; in our experiments we considered bottlenecks of 512 and 1024 neurons.
The decoder uses four convolution blocks with channels symmetrical to the encoder and interleaved by $2\times$ bilinear upsampling.
We attach to the bottleneck a convolution head responsible for localizing the drone, consisting of two convolution blocks with 32 and 1 channels and interleaved by $2\times$ bilinear upsampling.
 
\ac{clip} uses the pre-trained image encoder of the homonymous model~\cite{radford2021learning} as a feature extractor; specifically, we adopt the variant using the vision transformer ViT-B/32 
producing 512 features.
We follow a similar approach to what is presented in~\cite[Section 3]{radford2021learning} but keeping CLIP parameters frozen and replacing the logistic regression with a \ac{fc} head.
A \ac{fc} head uses two blocks composed of a \ac{fc} layer, batch normalization and ReLU activation.
%
%
We conducted many trials to find performing architectures that use the 512 CLIP features; we report here the best two:
the top performer with 16 neurons and the second with 512, followed by the output block of 400 neurons reshaped into a $20 \times 20$ grid.

Finally, we compare \ac{ledp} with Frontnet~\cite{bonato2023ultra}, an approach that directly regresses the drone's coordinates.

All \ac{nn}s are trained using Adam~\cite{adam} as the optimizer, running for a total of 200 epochs.
We adopt a scheduler that divides the learning rate by a factor of 5 every 50 epochs, starting with a learning rate $\eta_{\, \text{start}} = 1e^{-2}$ and reaching a final learning rate $\eta_{\, \text{final}} = 8e^{-5}$.
In each epoch we randomly draw mini-batches of 64 examples from the two joined training sets and minimize the loss function described in Section~\ref{sec:model}.
Specifically, we minimize ${\cal L}_\text{pretext}$ (setting $\mathcal{L}_\text{task}$ to 0) for examples taken from ${\cal T}_u$ since there are no known labels, 
and the complete loss ${\cal L}$ when fed samples from ${\cal T}_\ell$.

%
Additionally, to increase the variability of the drone's visual appearance, we apply the following augmentations:
horizontal flip (50\% probability), random rotation (uniform $\pm\ang{9}$), random translation (uniform $\pm32$ pixels), and apply multi-frequency simplex noise.

\subsection{From Grid Map to Robot Position}\label{sec:setup:argmax}

\begin{figure}[h]
    \centering
    \includegraphics[width=\linewidth]{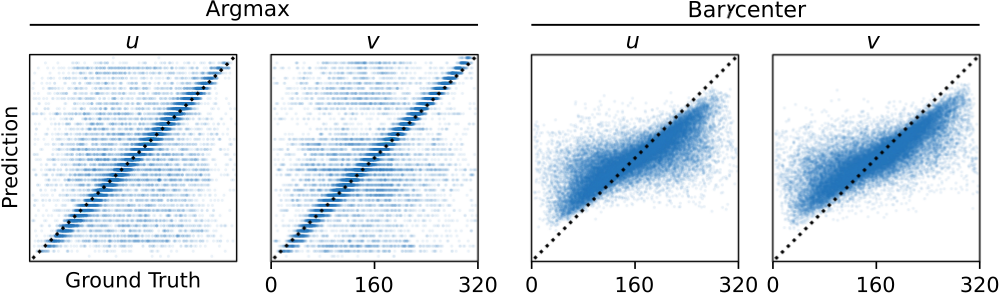}%
    \caption{\ac{ledp} model predictions on the test set $\mathcal{Q}$ with argmax and barycenter approaches for the $u$ and $v$ components of the drone's position.}
    \label{fig:scatter}
\end{figure}

\mdiff{We consider two approaches to recover $\hat{\bm{p}}$ from the model's predicted map $\hat{\bm{q}}$ named argmax and barycenter:
argmax selects the coordinate of the cell of $\hat{\bm{q}}$ whose value is largest;
barycenter computes the expected drone position by averaging the coordinates of each cell weighted by the corresponding probability of depicting a drone~\cite{finn2016deep}.
The probability of each cell is obtained by normalizing $\hat{\bm{q}}$ s.t. its sum equals one.
%
As shown in Figure~\ref{fig:scatter}, barycenter returns conservative estimates, biased towards the dataset's mean.}
In contrast, argmax yields unbiased results at the expense of larger errors for frames with no detected drone.
%
%
Banding artifacts are present with argmax since it cannot represent positions inbetween cells of the $40 \times 40$ map, i.e., it discretizes the input image coordinates into $8$-pixel-wide bins.
In the following experiments we use the argmax approach.

\subsection{Evaluation Metrics}

We compare models on different metrics computed on the test set $\mathcal{Q}$.
The Pearson correlation coefficient $\rho$, computed separately for the horizontal $u$ and vertical $v$ components of $\bm{p}$;
it measures the linear correlation between predicted and ground truth values.
We also compute the model's error distribution using the euclidean distance between $\bm{p}$ and $\hat{\bm{p}}$.
From this distribution, we derive the median value in pixels $\widetilde{\text{D}}$ and a precision score $\text{P}_k$.
We chose median instead of mean for being a robust central tendency estimate for skewed distributions.
$\text{P}_k$ is the fraction of samples whose position error is lower than $k$ pixels, considering predictions with a distance smaller than the threshold $k$ as correct, similarly to ADD for 6 Degrees of Freedom pose estimation~\cite{xiang2018posecnn}.
Additionally, we report $\text{P}^{+}_{k}$, defined as the relative improvement of \ac{ledp} models with respect to the corresponding baseline \ac{bas}, such that \ac{bas} represents 0\%, and \ac{ub} 100\%.
In our experiments we consider $k = 30$ pixels, i.e., approximately 10\% of the edge length of images.

Even though LED state prediction is not our end task, we also report the \ac{auc} for the LED classification output:
it measures how well a model distinguishes the two classes at various thresholds, i.e., telling between a drone with LEDs off or on.
A random classifier achieves an \ac{auc} score of 50\%, while an ideal classifier achieves a score of 100\%.

\section{Results and Discussion}\label{sec:results}

This section reports
the performance of our proposed strategy (Section~\ref{sec:results:performance}),
how performance changes as a function of $\lambda$ and of the amount of labeled examples (Section~\ref{sec:results:sensitivity}),
a comparison with alternative approaches (Section~\ref{sec:results:alternative}),
generalization ability of our proposed strategy (Section~\ref{sec:results:generalization})
and a deployed in-field experiment on nano-drones (Section~\ref{sec:results:infield}).

\begin{table*}[th]
    \setlength\tabcolsep{1.5mm} 
    \centering
    \caption{Comparison of models, 5 replicas per row. Pearson correlation coefficient $\rho_{u}$ and $\rho_{v}$, precision $\text{P}_{30}$ and median of the error $\widetilde{\text{D}}$.
    }
    \label{tab:thetable}
    \begin{tabular}{lcclrrrrr>{\centering\arraybackslash}p{6cm}}
    \toprule
    \\[-3.3mm]
    \multirow{2}{*}{Model} & \multicolumn{2}{c}{Training set for task} & \hfil\multirow{2}{*}{$\lambda$}\hfil & \multicolumn{1}{c}{$\rho_{u}$} & \multicolumn{1}{c}{$\rho_{v}$} & \multicolumn{1}{c}{$\text{P}_{30}$} & \multicolumn{1}{c}{$\text{P}^{+}_{30}$} & \multicolumn{1}{c}{$\widetilde{\text{D}}$} & 
    {Point plot for $\text{P}_{30}$ [\%] $\rightarrow$} \\
    & End & Pretext & & [\%] $\uparrow$ & [\%] $\uparrow$ & [\%] $\uparrow$ & [\%] $\uparrow$ & [px] $\downarrow$ & Error bars denote 95\% conf. int. \\
    \midrule
    DUMMY            & $-$ & $-$                                 & $-$ &  \multicolumn{1}{c}{$-$} &  \multicolumn{1}{c}{$-$} &  8.0 & $-$ &  79.0 & \multirow{1}{*}{\definecolor{wong_gray}{HTML}{888888}%
\definecolor{wong_black}{HTML}{333333}%
\definecolor{wong_gold}{HTML}{E69F00}%
\definecolor{wong_cyan}{HTML}{56B4E9}%
\definecolor{wong_green}{HTML}{009E73}%
\definecolor{wong_yellow}{HTML}{F0E442}%
\definecolor{wong_blue}{HTML}{0072B2}%
\definecolor{wong_red}{HTML}{D55E00}%
\definecolor{wong_pink}{HTML}{CC79A7}%
\definecolor{wong_magenta}{HTML}{CA1963}%
\begin{tikzpicture}[scale=0.0588]%
  \draw [opacity=0.0] (0, 0) -- (0, -145) -- (100, -145) -- (100, 0) -- cycle;
  \def\tablescatterdist{{8.0,63.4,80.9,84.6,86.9,85.2,57.5,60.4,76.2,70.5,68.3,60.9,39.9,50.3,43.8,42.5,3.7,3.8,6.5,5.7,48.7,73.3,93.9}}
  \def\tablescatterdiststd{{0.00,8.99,0.20,0.40,0.76,0.91,8.06,4.99,3.16,8.48,2.41,7.15,7.91,10.42,12.48,11.54,3.48,4.56,0.45,0.42,8.73,3.09,0.34}}
  \def\tablescatteradj{{0.0,0.0,0.0,0.0,0.0,0.0,2.9,2.9,2.9,2.9,2.9,5.8,5.8,5.8,5.8,5.8,8.5,8.5,8.5,8.5,8.5,8.5,11.4}}
  \def\tablescattercol{{"gray","green","green","green","green","gold","green","green","green","green","gold","green","green","green","green","gold","pink","pink","blue","blue","cyan","black","red"}}
  \def\confval{0.95}
  \newcommand{\tabscatx}[1]{\tablescatterdist[#1]}
  \newcommand{\tabscatxm}[1]{{\tablescatterdist[#1] - \confval * 0.447 * \tablescatterdiststd[#1]}}
  \newcommand{\tabscatxM}[1]{{\tablescatterdist[#1] + \confval * 0.447 * \tablescatterdiststd[#1]}}
  \newcommand{\tabscaty}[1]{{-5.4 * #1 - 2.1 - \tablescatteradj[#1]}}
  \newcommand{\tabscatc}[1]{\tablescattercol[#1]}
  \newcommand{\errorbar}[1]{\draw (\tabscatxm{#1}, \tabscaty{#1}) -- (\tabscatxM{#1}, \tabscaty{#1});\draw ($ (\tabscatxm{#1}, \tabscaty{#1}) + (0, -1.3) $) -- ($ (\tabscatxm{#1}, \tabscaty{#1}) + (0, +1.3) $);\draw ($ (\tabscatxM{#1}, \tabscaty{#1}) + (0, -1.3) $) -- ($ (\tabscatxM{#1}, \tabscaty{#1}) + (0, +1.3) $);}
  \newcommand{\pvaluel}[4][0]{\pgfmathsetmacro{\xM}{min(\tabscatxm{#2},\tabscatxm{#3})};\draw [gray] ($ (\tabscatxm{#2}, \tabscaty{#2}) + (-2, 0) $) -- ($ (\xM, \tabscaty{#2}) + ({-6 - #1}, 0) $) -- node[left] {\scriptsize{#4}} ($ (\xM, \tabscaty{#3}) + ({-6 - #1}, 0) $) -- ($ (\tabscatxm{#3}, \tabscaty{#3}) + (-2, 0) $);}
  \newcommand{\pvaluer}[4][0]{\pgfmathsetmacro{\xM}{max(\tabscatxM{#2},\tabscatxM{#3})};\draw [gray] ($ (\tabscatxM{#2}, \tabscaty{#2}) + (+2, 0) $) -- ($ (\xM, \tabscaty{#2}) + ({+6 + #1}, 0) $) -- node[right] {\scriptsize{#4}} ($ (\xM, \tabscaty{#3}) + ({+6 + #1}, 0) $) -- ($ (\tabscatxM{#3}, \tabscaty{#3}) + (+2, 0) $);}
  \newcommand{\vertbar}[4][0.5]{\pgfmathsetmacro{\c}{\tabscatc{#4}};\fill [{wong_\c}, opacity=0.3] ($ (\tabscatx{#4}, \tabscaty{#2}) - (#1, -2.5) $) rectangle ($ (\tabscatx{#4}, \tabscaty{#3}) + (#1, -2.5) $);}
  \vertbar{0}{5}{5}
  \vertbar{6}{10}{10}
  \vertbar{11}{15}{15}
  \vertbar{0}{22}{22}
  \foreach \i [evaluate=\i as \c using {\tabscatc{\i}}] in {0, ..., 22}{
    \errorbar{\i};
    \fill [{wong_\c}, radius=12mm, opacity=1.0] (\tabscatx{\i}, \tabscaty{\i}) circle;
  }
  \pvaluel{4}{5}{p=0.029}
  \pvaluer{8}{10}{p=0.014}
  \pvaluer{13}{15}{n.s.}
  \foreach \i in {0, 20, ..., 100}{
    \draw [font=\scriptsize] (\i, -139) -- ++(0, 1.5) node[draw=none, above]{\i};
  }
\end{tikzpicture}%
} \\
    \ac{ledp}-100    & ${\cal T}_\ell$ & ${\cal T}_\ell \cup {\cal T}_u$             & 0.0100 & 55.2 & 59.0 & 63.4 &-250.6 &  13.2 & \\
    \ac{ledp}-100    & ${\cal T}_\ell$ & ${\cal T}_\ell \cup {\cal T}_u$             & 0.0050 & 74.3 & 78.0 & 80.9 & -49.4 &   8.8 & \\
    \ac{ledp}-100    & ${\cal T}_\ell$ & ${\cal T}_\ell \cup {\cal T}_u$             & 0.0010 & 79.6 & 81.7 & 84.6 &  -6.9 &   8.3 & \\
    \ac{ledp}-100    & ${\cal T}_\ell$ & ${\cal T}_\ell \cup {\cal T}_u$             & 0.0005 & \textbf{81.3} & \textbf{83.6} & \textbf{86.9} &  \textbf{19.5} &   \textbf{8.0} & \\
    \ac{bas}-100     & ${\cal T}_\ell$ & $-$                                         & 0.0000 & 79.0 & 82.7 & 85.2 &   0.0 &   8.2 & \\
    \cmidrule{1-9}
    \ac{ledp}-30     & $30\%{\cal T}_\ell$ & $30\%{\cal T}_\ell \cup {\cal T}_u$     & 0.0100 & 50.1 & 55.5 & 57.5 & -43.0 &  14.9 & \\
    \ac{ledp}-30     & $30\%{\cal T}_\ell$ & $30\%{\cal T}_\ell \cup {\cal T}_u$     & 0.0050 & 51.0 & 57.3 & 60.4 & -30.8 &  14.3 & \\
    \ac{ledp}-30     & $30\%{\cal T}_\ell$ & $30\%{\cal T}_\ell \cup {\cal T}_u$     & 0.0010 & \textbf{68.5} & \textbf{71.4} & \textbf{76.2} &  \textbf{30.9} &   \textbf{9.9} & \\
    \ac{ledp}-30     & $30\%{\cal T}_\ell$ & $30\%{\cal T}_\ell \cup {\cal T}_u$     & 0.0005 & 61.7 & 66.3 & 70.5 &   8.6 &  11.1 & \\
    \ac{bas}-30      & $30\%{\cal T}_\ell$ & $-$                                     & 0.0000 & 59.0 & 66.2 & 68.3 &   0.0 &  10.9 & \\
    \cmidrule{1-9}
    \ac{ledp}-10     & $10\%{\cal T}_\ell$ & $10\%{\cal T}_\ell \cup {\cal T}_u$     & 0.0100 & 18.3 & 28.7 & 25.1 & -33.9 &  94.1 & \\
    \ac{ledp}-10     & $10\%{\cal T}_\ell$ & $10\%{\cal T}_\ell \cup {\cal T}_u$     & 0.0050 & 32.3 & 42.2 & 39.9 &  -5.1 &  61.0 & \\
    \ac{ledp}-10     & $10\%{\cal T}_\ell$ & $10\%{\cal T}_\ell \cup {\cal T}_u$     & 0.0010 & \textbf{42.1} & \textbf{48.0} & \textbf{50.3} &  \textbf{15.2} &  \textbf{33.6} & \\
    \ac{ledp}-10     & $10\%{\cal T}_\ell$ & $10\%{\cal T}_\ell \cup {\cal T}_u$     & 0.0005 & 36.1 & 44.7 & 43.8 &   2.5 &  53.1 & \\
    \ac{bas}-10      & $10\%{\cal T}_\ell$ & $-$                                     & 0.0000 & 34.7 & 41.6 & 42.5 &   0.0 &  57.8 & \\
    \cmidrule{1-9}
    \ac{ae}-512      & ${\cal T}_\ell$ & ${\cal T}_\ell \cup {\cal T}_u$             & $-$    &  0.5 &  1.1 &  3.7 &   $-$ & 137.9 & \\
    \ac{ae}-1024     & ${\cal T}_\ell$ & ${\cal T}_\ell \cup {\cal T}_u$             & $-$    & -0.9 & -2.2 &  3.8 &   $-$ & 130.1 & \\
    \ac{clip}-16~\cite{radford2021learning}      & ${\cal T}_\ell$ & $-$             & $-$    &  2.0 &  8.9 &  6.5 &   $-$ & 102.7 & \\
    \ac{clip}-512~\cite{radford2021learning}     & ${\cal T}_\ell$ & $-$             & $-$    &  1.4 &  7.8 &  5.7 &   $-$ & 109.1 & \\
    Frontnet~\cite{bonato2023ultra}              & ${\cal T}_\ell$ & $-$             & $-$    & \textbf{69.5} & \textbf{74.1} & 48.7 &   $-$ &  31.0 & \\
    \ac{decro}~\cite{li2022self}                 & ${\cal T}_\ell$ & ${\cal T}_s$    & $-$    & 64.7 & 68.5 & \textbf{73.3} &   $-$ &  \textbf{10.2} & \\
    \cmidrule{1-9}
    \ac{ub}          & ${\cal T}_\ell \cup {\cal T}_u^*$ & $-$                       & 0.0000 & 91.7 & 89.3 & 93.9 & 100.0 &   6.8 &
    \\[1.3mm]
    \bottomrule
    \end{tabular}
\end{table*}

\subsection{LED State Prediction Improves Performance}\label{sec:results:performance}

In Table~\ref{tab:thetable}, first and last panels, we report the performance of our \ac{ledp} strategy against a dummy model (DUMMY), baseline (\ac{bas}) and an upperbound (\ac{ub}).
We observe that \ac{ledp}-100, trained leveraging unlabeled images, performs moderately better than \ac{bas}-100 across all evaluation metrics.
The $\text{P}_{30}$ metric indicates that 86.9\% of predictions fall within 30 pixels from the respective ground truth position, with a median error of only 8 pixels out of a $320 \times 320$ pixels image.
We also computed $\text{P}_{30}$ scores on two subsets of $\mathcal{Q}$ containing examples with LEDs off and on, on which our \ac{ledp}-100 model scores 83.3\% and 91.2\% respectively, 
showing more difficulties in localizing drones with LEDs off.
On the same metric, we report \ac{ledp}-100 to score higher than \ac{bas}-100 with a p-value of 0.029, computed with the non-parametric one-sided Mann-Whitney U test.
%
\mdiff{This model achieves a $\text{P}_{30}$ of 88.3\% when localizing the drone in brighter images and 85.3\% in darker ones, showing a slight performance drop in the latter case.}

\subsection{Impact of \texorpdfstring{$\lambda$}{Lambda} and amount of Labeled Examples}\label{sec:results:sensitivity}

\begin{figure*}[th]
    \centering
    \includegraphics[width=\linewidth]{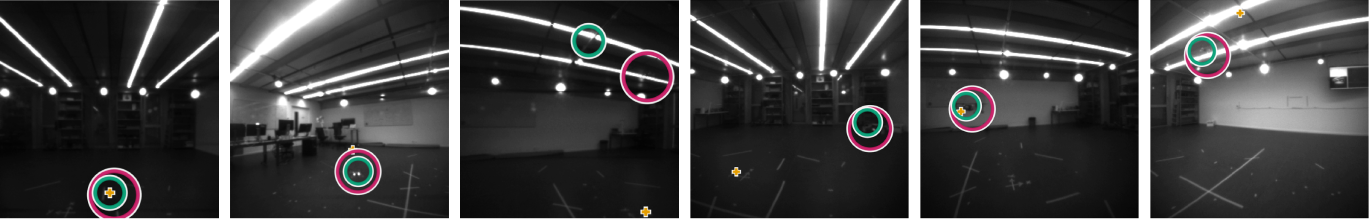}%
    \caption{\mdiff{\ac{ledp}-30 with $\lambda = 0.001$ (small green circle), \ac{bas}-30 (yellow cross) predictions and ground truth (large magenta circle) on frames taken from $\mathcal{Q}$ with the drone's LEDs turned on (first three) and off (last three), and featuring different camera exposure settings}.}
    \label{fig:predict}
\end{figure*}

\mdiff{In Figure~\ref{fig:predict}, we inspect the \ac{ledp}-30 ($\lambda = 0.001$) prediction against \ac{bas}-30 on samples taken from $\mathcal{Q}$,
where it scores 30.9\% in $\text{P}^{+}_{30}$, a considerable improvement over the respective baseline.
Our model demonstrates an overall good performance when the target drone flies at or below the camera height as the drone's LEDs are more easily visible, and to a lesser extent when the target flies higher, which reduces the LEDs' visibility.
%
%
On the LED state prediction, our model scores 74\% in AUC despite the drone LEDs not being visible in many of $\mathcal{Q}$ samples.
Failed detections occur when the model predicts the position of similar looking areas of the image;
this happens less frequently when LEDs are on since their presence improves the drone's visibility.

The approach is robust to the unlabeled training set $\mathcal{T}_u$ containing some frames in which the drone is out of the field of view, thus making its LED state impossible to predict.
We explore the approach robustness in an experiment where we add 3.4k such samples to $\mathcal{T}_u$, a 10\% increase.
After training on the modified $\mathcal{T}_u$, LED-P-30 ($\lambda = 0.001$) scores 10.3 in $\widetilde{\text{D}}$ and 73.2\% in $\text{P}_{30}$, a small decline in performance when compared to the same model trained using the original $\mathcal{T}_u$.}

\begin{figure}[th]
    \centering
    \includegraphics[width=0.97\linewidth]{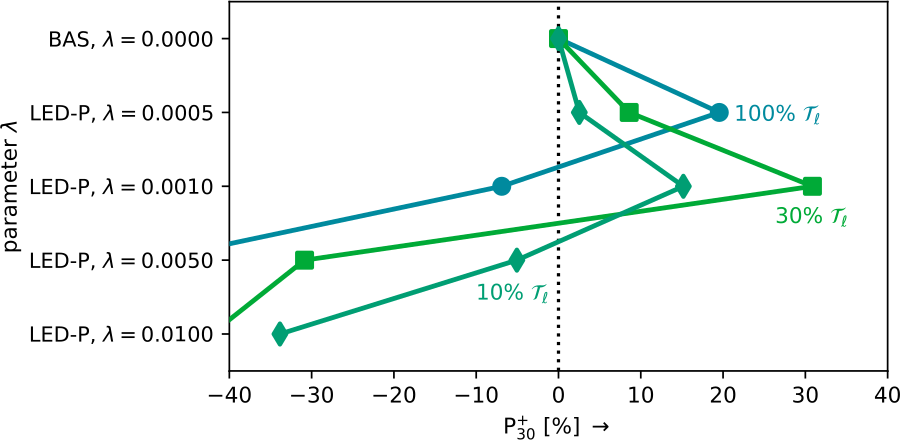}%
    \caption{$\text{P}^{+}_{30}$ score for \ac{bas} $(\lambda = 1)$ and \ac{ledp} $(\lambda < 1)$ strategies as the amount of labeled training examples $\mathcal{T}_\ell$ and the weight of the loss $\lambda$ vary.}
    \label{fig:sensitivity}
\end{figure}

In Figure~\ref{fig:sensitivity}, we show the $\text{P}^{+}_{30}$ relative improvement score for \ac{bas} and \ac{ledp} strategies as the amount of labeled training examples and the weight of the loss $\lambda$ vary.
\ac{ledp} outperforms \ac{bas} when training on as few as 100 labeled examples (10\% of $\mathcal{T}_\ell$).
The optimal value of $\lambda$ for our loss is 0.001 for 100 and 300 labeled examples (10\% and 30\% of $\mathcal{T}_\ell$ respectively), and 0.0005 for the full labeled dataset.

Inspecting the two loss term values, we note that $\mathcal{L}_\text{task}$ is in the order of magnitude of $10^{-4}$ and $\mathcal{L}_\text{pretext}$ in $10^{-1}$.
%
The optimal $\lambda$ values are scaling the loss terms to be in the same order of magnitude, striking a balance between the two.

\subsection{Alternative Training Strategies}\label{sec:results:alternative}

%
We investigate strategies using a different architecture, model pre-training, or different pretext tasks, described in Section~\ref{sec:setup:strategies}, and whose performance metrics are reported in Table~\ref{tab:thetable}, fourth panel.
\ac{decro} scores 10.2 pixels in $\widetilde{\text{D}}$ demonstrating some degree of accuracy; however, it scores lower than \ac{ledp}-100 and \ac{bas}-100 that use the same amount of labeled data.
This result suggests that task similarity is less influential than dataset relevance, i.e., training on the same task with data vastly different (in appearance) from testing achieves lower scores than solving a different task on similar data, e.g., our LED state prediction pretext task.

Frontnet achieves 31 pixels in $\widetilde{\text{D}}$ despite having been trained similarly to \ac{bas}-100, which achieves only 8.2 pixels.
The increase in error demonstrated by Frontnet indicates that using a \ac{fcn} model producing a map-based representation leads to a better performance than direct regression. 

\ac{ae} successfully learns to reconstruct input images using the bottleneck features.
However, the model focuses on large-scale aspects of the environment, such as floors, walls and fixtures, distinctive of background elements and  disregards high frequency elements such as the drone.
\mdiff{This tendency results in a feature space, regardless of bottleneck size, that is \emph{not informative} of the drone's position, rendering this pretext task inadequate for localizing small objects.
Even in \ac{clip}'s case, we note how the provided features do not translate in good localization performance;
%
%
this confirms previous reports~\cite{radford2021learning} that CLIP's image encoder features underperform on highly specialized end tasks, such as ours. 

\ac{ae} and \ac{clip} highlight the importance of having good features, which can be obtained by choosing the right kind of pretext task, and promoting the recognition of patterns similar to those required to solve the end task.}

\subsection{Generalization Ability}\label{sec:results:generalization}

\begin{figure*}[th]
    \centering
    \includegraphics[width=\linewidth]{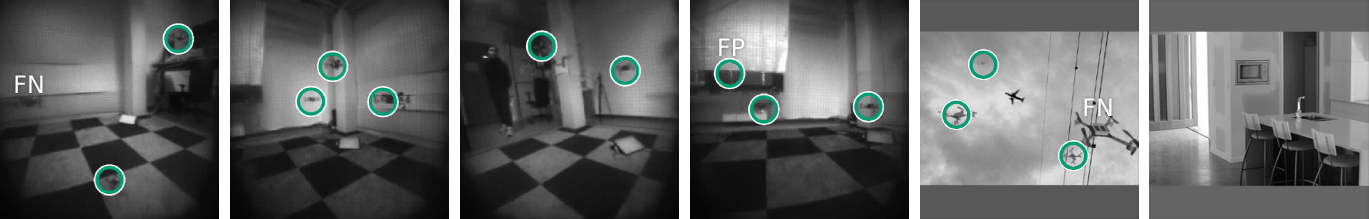}%
    \caption{\mdiff{Generalization and multi-drone localization examples using \ac{ledp}-30 ($\lambda = 0.001$). The first four images are taken from an unseen lab environment~\cite{moldagalieva2023dataset} and the last two from an unseen synthetic one~\cite{li2022self}. Errors are visually marked as false positives (FP) or false negatives (FN).}}
    \label{fig:multi}
\end{figure*}

\mdiff{In Figure~\ref{fig:multi}, we show the prediction of the observer drone using \ac{ledp}-30 ($\lambda = 0.001$) on images featuring multiple target drones collected in another lab environment~\cite{moldagalieva2023dataset}, and synthetic frames from a simulator~\cite{li2022self}.
For this scenario, we modified the argmax approach by thresholding the localization map $\hat{\bm{l}}$ with its 95-percentile, extracting the maximum value of each connected component, discarding components whose maximum is below 0.2 and returning their centroid as the drone locations.
For the most part, our model correctly localizes the drones despite the motion blur and defocus, with all examples featuring at least two correct predictions.
Failed detections on the edge of the field of view occur due a strong vignetting effect, which degrades the image quality.}

\subsection{In-field Experiment}\label{sec:results:infield}

\mdiff{The \ac{ledp}-30 ($\lambda = 0.001$) and \ac{bas}-30 models are deployed aboard the observer nano-drone, using the academic NEMO/DORY framework\footnote{\href{https://github.com/pulp-platform/nemo}{https://github.com/pulp-platform/nemo}}.
NEMO provides post-training quantization-aware fine-tuning, to convert deep learning models from floating-point to integer arithmetic, needed due to the absence of floating point units on the GAP8 \ac{soc}.
DORY, instead, produces a template-based C implementation, which takes care of data movements across the memory hierarchy of the GAP8 \ac{soc}.
This stage is fundamental in achieving fast inference, as sub-optimal data tiling/transfers might lead to poor performances.
Our optimized NN pipeline achieves an in-field inference rate of 21 frames per second.}

In the observer drone, the model output is used as feedback to a visual servoing controller, designed to keep the target drone in the center of the image. 
The controller moves the observer drone on a vertical plane, orthogonal to its camera axis, while keeping a constant yaw.  
Without loss of generality, we consider the motion of the observer drone to take place on the $yz$ world plane, with $x=0$.
In the experiment, the target drone follows a predefined, scripted trajectory.
The ideal trajectory for the observer drone is the same as the target, projected on $x=0$ vertical world plane.

\begin{figure}[th]
    \centering
    \includegraphics[width=0.90\linewidth]{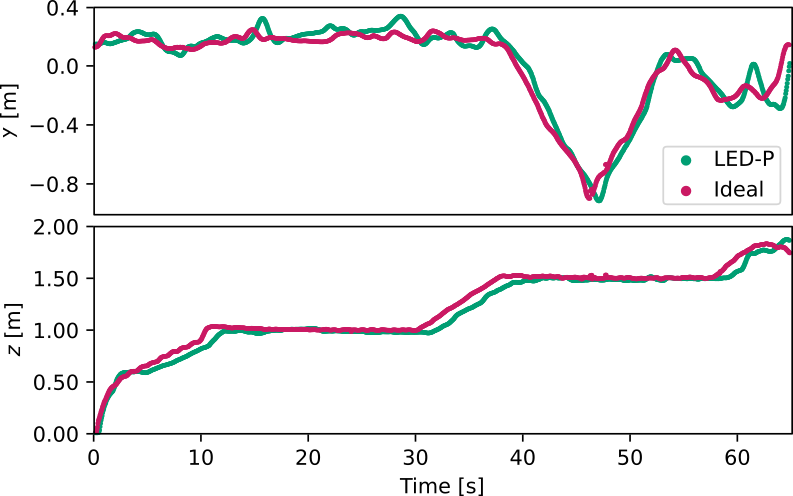}%
    \caption{In-field experiment: measured vs ideal trajectory of the observer drone when using \ac{ledp}-30 ($\lambda = 0.001$) for estimating the target position.}
    \label{fig:infield}
\end{figure}

Figure~\ref{fig:infield} reports the $y$ and $z$ components of the measured trajectory of the observer drone, controlled using position estimates from \ac{ledp}-30, compared to the ideal one.
We observe that the drone follows very closely the ideal trajectory.
The same experiment run using \ac{bas}-30 yields worse position tracking: the mean and standard deviation $\sigma$ of the absolute position error on the $yz$ plane is 4.2 cm ($\sigma=4.0$ cm) for \ac{ledp}-30, and 11.9 cm ($\sigma=8.3$ cm) for \ac{bas}-30.

\section{Conclusions}\label{sec:conclusions}

We propose LED state estimation as a self-supervised pretext task, applied to the end task of visually localizing robots from small labeled datasets.
The pretext task is optimized on large, cheaply-collected datasets that only have ground truth for the LED state of the observed robot.
The approach is instantiated on localizing nano-quadrotors in low-resolution images, observing improved localization accuracy compared to baselines and alternative techniques for self-supervision.
In-field experiments used a 27-g Crazyflie nano-drone to track the position of a peer drone; the proposed approach reduces mean position-tracking error from 11.9 to 4.2 cm.
\mdiff{The resulting detector can be used, for example, within a tracking-by-detection approach~\cite{tian2018detection} to integrate predictions over time and track the target even in presence of occlusions.

Current work aims at extending the approach to estimate the distance and orientation of the target, handling sequential inputs, and exploiting the known state of multiple LEDs as a localization cue at inference time.}


\bstctlcite{IEEEexample:BSTcontrol}

\bibliographystyle{IEEEtran}
\bibliography{references}

\end{document}